# Finding Common Characteristics Among NBA Playoff and Championship Teams: A Machine Learning Approach

By: Dr. Ikjyot Singh Kohli
Department of Mathematics and Statistics – York University – Toronto, Ontario
e-mail: isk@mathstat.yorku.ca

April 3, 2017

## Abstract

In this paper, we employ machine learning techniques to analyze seventeen seasons (1999-2000 to 2015-2016) of NBA regular season data from every team to determine the common characteristics among NBA playoff teams. Each team was characterized by 26 predictor variables and one binary response variable taking on a value of "TRUE" if a team had made the playoffs, and value of "FALSE" if a team had missed the playoffs. After fitting an initial classification tree to this problem, this tree was then pruned which decreased the test error rate. Further to this, a random forest of classification trees was grown which provided a very accurate model from which a variable importance plot was generated to determine which predictor variables had the greatest influence on the response variable. The result of this work was the conclusion that the most important factors in characterizing a team's playoff eligibility are a team's opponent number of assists per game, a team's opponent number of made two point shots per game, and a team's number of steals per game. This seems to suggest that defensive factors as opposed to offensive factors are the most important characteristics shared among NBA playoff teams. We then use neural networks to classify championship teams based on regular season data. From this, we show that the most important factor in a team not winning a championship is that team's opponent number of made three-point shots per game. This once again implies that defensive characteristics are of great importance in not only determining a team's playoff eligibility, but certainly, one can conclude that a lack of perimeter defense negatively impacts a team's championship chances in a given season. Further, it is shown that made two-point shots and defensive rebounding are by far the most important factor in a team's chances at winning a championship in a given season.

## 1. Introduction

Can one based on data science methodologies effectively predict which NBA teams will make the playoffs and/or win a championship in a given year? This is the question that we have chosen to analyze in this paper. The use of data science methodologies, popularly and broadly termed as "analytics" has been on an increase over the last number of years. Some examples can be found in [1, 2, 3, 4, 5, 6, 7] and references therein.

To look at this problem in particular, we analyzed seventeen years of NBA team regular season data, that is, from the 1999-2000 NBA Season to the 2015-2016 NBA Season using the data available from Basketball-Reference.com [8]. We generated a dataset that associated with each NBA team, 26 predictor variables that classified a given team's performance during the regular season. We then generated a binary response variable that took the value of "TRUE" if a certain team made the playoffs, and "FALSE" if a certain team missed the playoffs. A sample of this dataset is provided in Table 1 below.

| X3P | X3PA | X2P | X2PA | FT | FTA | Playoffs |
|---|---|---|---|---|---|---|
| 3.146341463 | 9.926829268 | 33.43902439 | 73.08536585 | 18.01219512 | 24.23170732 | FALSE |
| 5.085365854 | 15.36585366 | 32.15853659 | 68.53658537 | 19.76829268 | 26.52439024 | FALSE |
| 4.134146341 | 12.20731707 | 31.65853659 | 67.46341463 | 22.7195122 | 29.97560976 | TRUE |
| 4.146341463 | 12.58536585 | 27.13414634 | 62.7804878 | 18.07317073 | 25.47560976 | FALSE |
| 4.182926829 | 11.20731707 | 32.12195122 | 70.91463415 | 20.15853659 | 26.8902439 | FALSE |
| 6.329268293 | 16.17073171 | 32.63414634 | 69.76829268 | 17.15853659 | 21.35365854 | FALSE |
| 5.731707317 | 17.03658537 | 31.54878049 | 67.24390244 | 18.67073171 | 25.80487805 | TRUE |

Table 1: A sample of our dataset.

Since the symbols in the column headers of Table 1 in addition to other symbols not displayed in Table 1 appear throughout the paper, their meanings are as follows: X3P = Three-point shots made per game, X3PA = Three-point shots attempted per game, X2P = Two-point shots made per game, X2PA = Two-point shots attempted per game, FT = Free throws made per game, FTA = Free throws attempted per game, ORB = Offensive rebounds per game, DRB = Defensive rebounds per game, AST = Assists per game, STL = Steals per game, BLK = blocks per game, TOV = turnovers per game, PF = personal fouls per game. Also, note that a small letter 'o' preceding each variable indicates the same statistic for a given team's opponent. Therefore, altogether, we considered a total of 26 predictor variables to characterize each team's performance during the season. To simplify our dataset and inputs into our machine learning algorithms below we intentionally did not consider predictor variables that were dependent on these "basic" variables. So, for example, we did not consider a team's field goal percentage, because this is simply a team's FG/FGA.

Given that our response variable is qualitative, it seems that the problem at hand is ripe for an analysis using classification trees. Following [9, 10], we use recursive binary splitting to grow our classification tree, using the minimum Gini index as the criterion for making the binary splits. Specifically, let $\hat{p}_{mk}$ denote the proportion of training observations in the $m^{th}$ region that are from the $k^{th}$ class. One then defines the Gini index as

$$G = \sum_{k=1}^{K} \hat{p}_{mk}(1 - \hat{p}_{mk})$$

where K represents the total number of classes. Note that G is relatively small if all of the $\hat{p}_{mk}$'s are close to zero or one. One can then consider G as a measure of node purity. Therefore, building a classification tree involves making binary splits, which maximize the reduction in node impurity, or, in other words minimizing G, the Gini index.

## 2 Fitting Classification Trees

In our first attempt at analyzing the question at hand, we split the original data set into a training set and a test set, in a 60% - 40% proportion. We then fit a classification tree on the training data, of which the result is shown in Fig. 1.

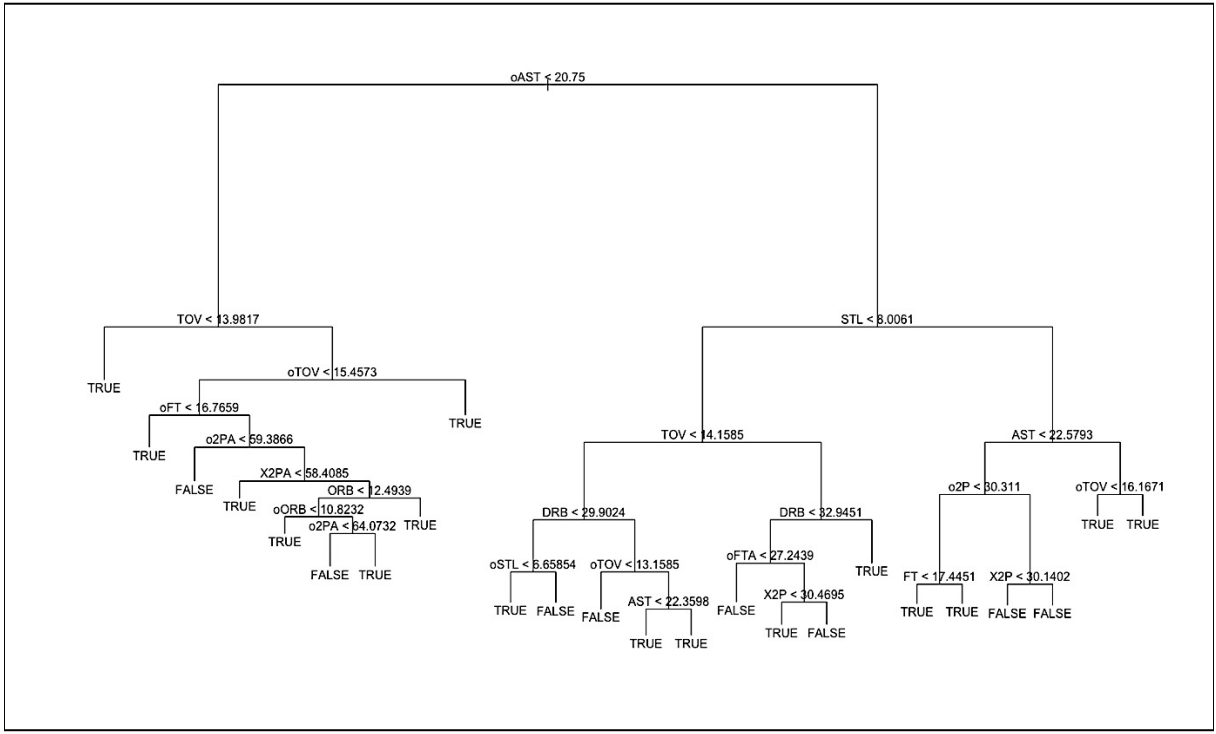

Figure 1: The initial classification tree fit to the training data categorizing a team's playoff success.

To evaluate the performance of this classification tree, we generated the confusion matrix as described in Table 2. One can see from this confusion matrix that (61+92)/202=75.74% of the test observations were classified correctly.

|       | FALSE | TRUE |
|------:|------:|-----:|
| FALSE |    61 |   18 |
|  TRUE |    31 |   92 |

Table 2: The confusion matrix created from fitting a classification tree to the training data and making predictions using the test data.

We now consider whether pruning the tree will have any effect on the classification performance. We wish to consider pruning as the original classification tree displayed in Fig. 1 may be too complex resulting from an overfit on the training dataset. A smaller tree with fewer splits may lead to a less-biased result [9]. To accomplish this, we first performed a cross-validation to determine the optimal level of tree complexity, and then used cost-complexity pruning to select an appropriate sequence of trees. Using the cv.tree() function in R, we found that the lowest cross validation error rate corresponded to a pruned tree with 8 terminal nodes. This classification tree is displayed in Fig. 2.

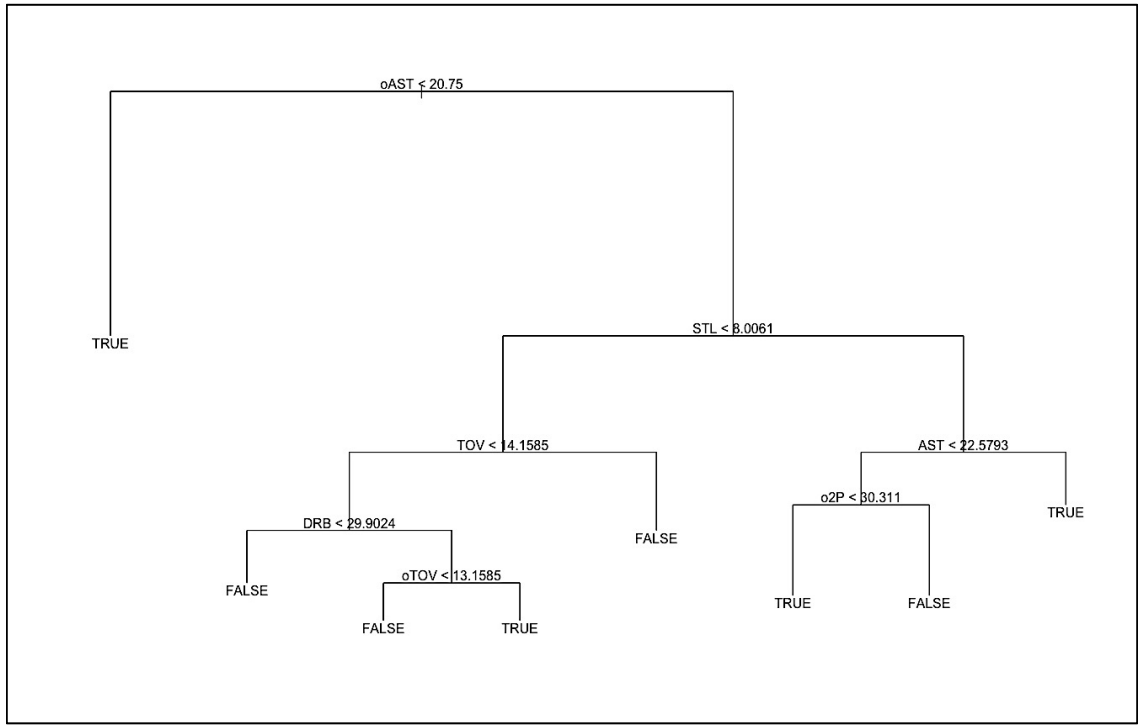

Figure 2: The pruned tree categorizing a team's playoff success. The pruned tree has 8 terminal nodes as opposed to 24 terminal nodes of the unpruned tree in Figure 1.

We now evaluate the performance of this pruned classification tree, by using the same splitting of test and training data as before, which generated the confusion matrix as described in Table 3. One can see from this confusion matrix that (66+87)/202=75.74% of the test observations were classified correctly.

|  | FALSE | TRUE |
|---|---|---|
| FALSE | 66 | 23 |
| TRUE | 26 | 87 |

Table 3: The confusion matrix created from fitting a pruned classification tree to the training data and making predictions using the test data, which consisted of 202 observations.

From the pruned tree in Fig. 2, we notice several "paths" for a team to make the playoffs. Namely, we have that:

$oAST < 20.75,$

$oAST > 20.75 \rightarrow STL > 8.0061 \rightarrow AST > 22.5793,$

$oAST > 20.75 \rightarrow STL > 8.0061 \rightarrow AST < 22.5793, o2P < 30.311,$

$oAST > 20.75 \rightarrow STL < 8.0061 \rightarrow TOV < 14.1585 \rightarrow DRB > 29.9024 \rightarrow oTOV > 13.1585.$

One sees that each "path" to the playoffs begins with a defensive factor, namely the opposing team's number of assists per game. Thus, the ability to disrupt a team's offensive flow by limiting the number of assists it has per game is evidently quite important. The alternative paths indicate a high number of steals per game, which is again a defensive factor, followed by combinations of numbers of assists, limiting the opposing team's number of two-point shots, a high number of defensive rebounds, and/or forcing turnovers. One sees that defensive factors are largely more important in determining whether a team makes the playoffs as opposed to offensive factors.

We finally build a random forest of classification trees in an attempt to construct a more powerful prediction model. Following [9], we note that this involves building a number of classification trees on bootstrapped training samples. Each time a split in a tree is considered, a random of sample of m predictors is chosen as split candidates from the full set of p predictors. The split itself is allowed to use only of the m predictors. The end result of this is that the trees will be decorrelated, making the average of the resulting trees less variable and more reliable.

Since there are 26 predictor variables, we consider $\sqrt{26} \sim 5$ variables randomly sampled as candidates at each split. The confusion matrix that was generated is displayed in Table 4. From this confusion matrix, we found that (76+95)/202=83.663% of the observations were classified correctly.

|  | FALSE | TRUE |
|---|---|---|
| FALSE | 74 | 15 |
| TRUE | 18 | 95 |

Table 4: The confusion matrix created from building a random forest of classification trees.

From this random forest of classification trees, we can generate a plot of the importance of each variable which is displayed in Fig. 3.

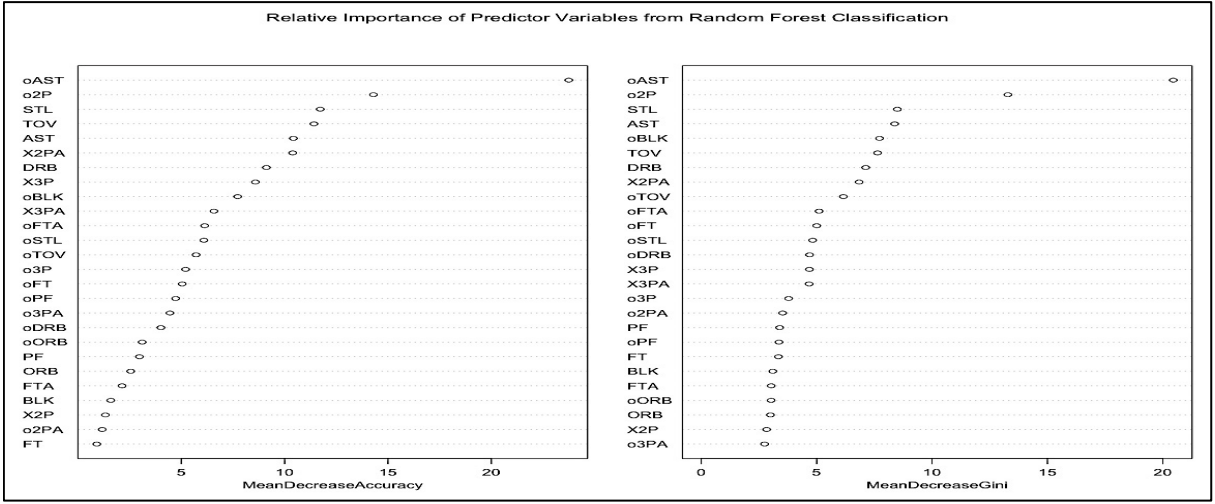

Figure 3: The importance of each predictor variable as computed from building the random forests of classification trees.

One sees that by far, the random forest employed was the most accurate as it classified 83.663% of the test observations correctly. More importantly, one sees from the importance plot in Fig. 3 that the predictor variables that have the most influence in deciding whether or not a team makes the playoffs are oAST (an opposing team's number of assists per game) and o2P (an opposing team's number of made two point shots per game). This approach once again affirms some of our previous results. Namely, defensive factors are much more important in deciding whether a team makes the playoffs as opposed to offensive factors.

## 3 Neural Networks and NBA Championship Teams

In this section, we amend the original dataset above by adding a column indicating whether a specific team had won an NBA Championship by defining a binary response variable to take the value of "TRUE" if this were true, and "FALSE" if this were false.

Following the arguments in [11], we use a multilayer perceptron (MLP) as a feedforward neural network. Each node in the MLP is termed a neuron, and inputs to the neurons in the first layer, which is the input layer, are the network inputs while outputs of the neurons in the last layer (output layer) are the network outputs. Layers between the input and the output layer are called hidden layers. In fact, let us denote by p the number of input neurons, and q, the number of output neurons. Further, denote by $w_{ij}$ the real-valued weight of each directed edge from neuron i to neuron j. Then, the effective input to each neuron is given by

$$I_i = w_i^j O_j,$$

where $O_j$ denotes the output of each neuron i. Note that we have employed the Einstein summation convention, where repeated indices are understood to be summed. One then relates the output of each neuron from the effective input via a nonlinear activation function, $s: \mathbb{R} \to \mathbb{R}$, so that $O_i = s(I_i)$. For our analysis, we use the logistic function as the activation function,

$$s(x) = \frac{1}{1 + e^{-x}} \in (0,1).$$

The main point is that this function maps the possibly infinite range of neuron inputs to a finite range of neuron outputs. The MLP then realizes a function $f: \mathbb{R}^p \to \mathbb{R}^q$ specified by the weights $w_{ij}$. Indeed, in a nonlinear regression problem, a set $Z = \{(x_1, y_1), \ldots, (x_n, y_n)\} \subset \mathbb{R}^{p+q}$ is used to estimate the weights $w_{ij}$ so that for each input vector $x_k$ the MLP estimates the corresponding output vector $y_k$. For a given neural network structure, the iterative method of gradient descent is used to find the weights $w_{ij}$. The interested reader is asked to consult [11] or [10] for further details.

To fit a neural network to our problem, we first used extensive cross-validation to find that the optimal neural network was one in which the decay parameter was $10^{-7}$ with 20 nodes in the hidden layer. Note that for this analysis, we used 60% of the data as training data and 40% of the data as a test data set. We used the nnet function [12] in R to accomplish this analysis. A plot of this neural network is given in Fig. 4 below accomplished via the plot.nnet function designed by Beck [13].

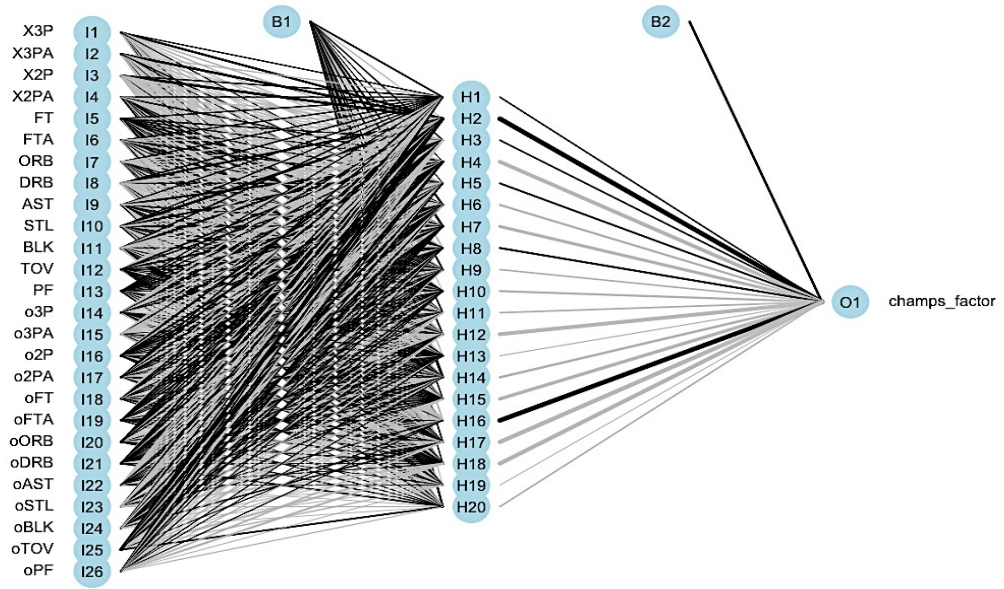

Figure 4: The neural network fitted to the classification problem in question. The black lines indicate positive

weights, while the grey lines indicate negative weights. The input layers are denoted by I1-I26, while the hidden layers are denoted by H1–H20, with the binary output layer being denoted by O1. Further, B1 and B2 indicate bias layers.

The evaluation of this neural network's performance on the test data is given by the confusion matrix generated in Table 5 below. One can see that the neural network classified 194/202=96.04% of the test observations correctly.

|  | FALSE | TRUE |
|---|---|---|
| FALSE | 193 | 4 |
| TRUE | 4 | 1 |

Table 5: The confusion matrix created from evaluating the neural network fit against the test data.

What is of special importance in such an analysis is to determine which of the predictor variables had the highest relevant importance on the response. For neural networks, we use a method proposed by Garson [14] and modified by Beck [15] to accomplish this task. The relative importance plot can be seen in Fig. 5 below.

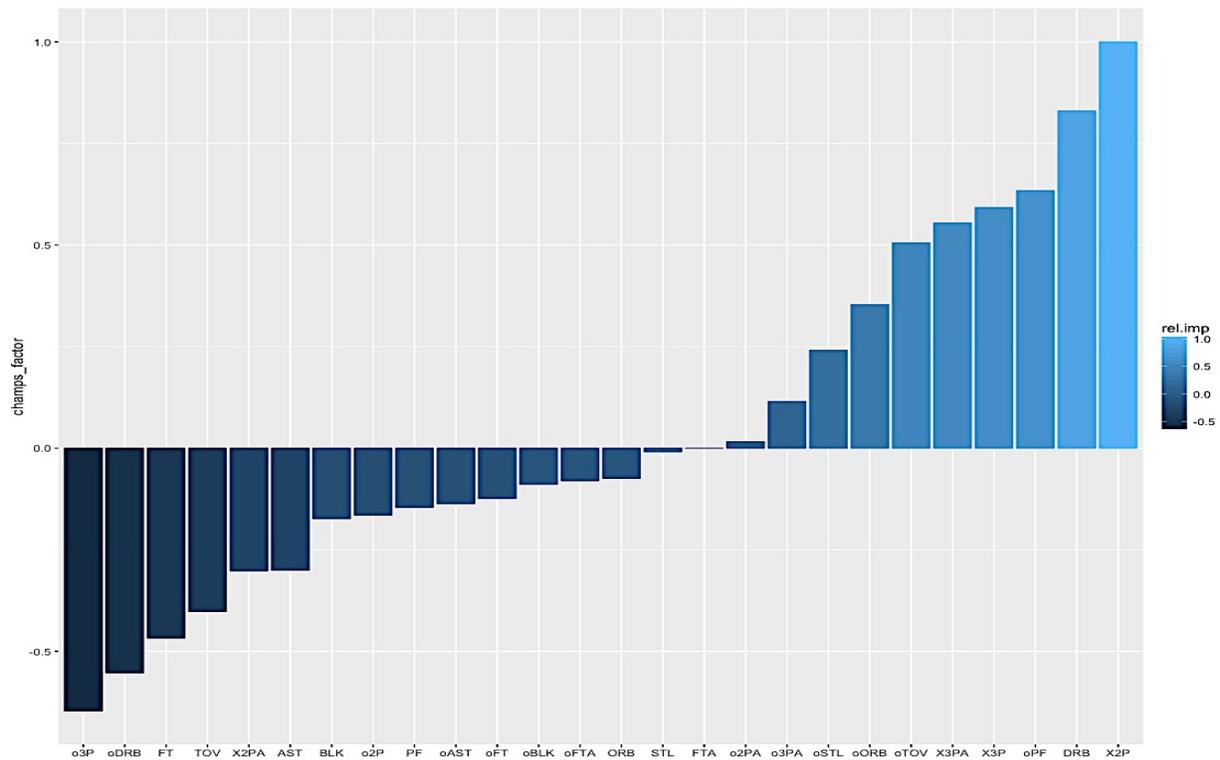

Figure 5: The relative importance plot corresponding to the neural network for classifying NBA championship teams.

From the importance plot in Fig. 5, one can therefore at least see that the most important factor in a team not winning a championship is that team's opponent number of made three-point shots per game. This once again implies that defensive characteristics are of great importance in not only determining a team's playoff eligibility, but certainly, one can conclude that a lack of perimeter defense negatively impacts a team's championship chances in a given season. Further, it is clear from this relative importance plot, that made two-point shots and defensive rebounding are by far the most important factors in a team's chances at winning a championship.

## 4 Conclusions

In this paper, we employed machine learning techniques to analyze seventeen seasons (1999-2000 to 2015-2016) of NBA regular season data from every team to determine the common characteristics among NBA playoff teams. Each team was characterized by 26 predictor variables and one binary response variable taking on a value of "TRUE" if a team had made the playoffs, and value of "FALSE" if a team had missed the playoffs. After fitting an initial classification tree to this problem, this tree was then pruned which decreased the test error rate. Further to this, a random forest of classification trees was grown which provided a very accurate model from which a variable importance plot was generated to determine which predictor variables had the greatest influence on the response variable. The result of this work was the conclusion that the most important factors in characterizing a team's playoff eligibility are a team's opponent number of assists per game, a team's opponent number of made two point shots per game, and a team's number of steals per game.  This seemed to suggest that defensive factors as opposed to offensive factors were the most important characteristics shared among NBA playoff teams. Further, from the importance plot in Fig. 5, one can therefore at least see that the most important factor in a team not winning a championship is that team's opponent number of made three-point shots per game. This once again implies that defensive characteristics are of great importance in not only determining a team's playoff eligibility, but certainly, one can conclude that a lack of perimeter defense negatively impacts a team's championship chances in a given season. Further, it is clear from this relative importance plot, that made two-point shots and defensive rebounding are by far the most important factors in a team's chances at winning a championship.

     Finally, this analysis will hopefully dispel the notion that has gained some momentum in recent years, that an offense geared towards attempting many three-point shots is a sufficient and necessary condition for an NBA team to be successful in qualifying for the playoffs and winning a championship as implied in [16] [17] [18] for example.